\documentclass[11pt,letterpaper]{article}
\usepackage[utf8]{inputenc}
\usepackage[margin=1in]{geometry}
\usepackage{times}
\usepackage{setspace}
\usepackage{graphicx}
\usepackage{amsmath}
\usepackage{amssymb}
\usepackage{cite}
\usepackage{hyperref}
\usepackage{booktabs}
\usepackage{multirow}
\usepackage{array}
\usepackage{caption}
\usepackage{fancyhdr}
\usepackage{float}

\usepackage{parskip}
\begin{document}

\begin{titlepage}
\centering
\vspace*{2in}
{\LARGE\bfseries Enhancing Neuro-Oncology Through Self-Assessing Deep Learning Models for Brain Tumor and Whole Brain Segmentation\par}
\vspace{2in}
{\large Andrew Zhou\par}
{\large The Quarry Lane School\par}
{\large Dublin, CA\par}
\vfill
\end{titlepage}
\thispagestyle{empty}

\section*{Abstract}
Accurate segmentation of brain tumors is critical for clinical diagnosis, surgical planning, and treatment monitoring. While deep learning models have achieved substantial progress on benchmark datasets, two key limitations hinder clinical adoption: the absence of uncertainty estimation for potential errors and the lack of healthy brain structure segmentation surrounding cancer needed for surgical planning. Current methods cannot provide clinicians with both tumor localization and surrounding anatomical context in a unified framework and lack a confidence estimate.

This study introduces an uncertainty-aware brain tumor segmentation framework that augments the nnU-Net model \cite{nnUnet} with an additional prediction channel for voxel-wise uncertainty estimation. The model was trained and evaluated on the BraTS2023 dataset \cite{BraTS2023} and achieves strong uncertainty prediction with a correlation coefficient of 0.634 and RMSD of 0.032 without sacrificing tumor segmentation accuracy. The proposed method predicts uncertainty directly from the deep learning model and therefore does not require multiple networks or multiple inference passes, which is a significant benefit over existing methods and enhances its potential utility in clinical decision-making. 

To further aid physicians in reviewing tumors in whole brain context, we build upon our previously published unified segmentation model that jointly segments tumor and whole brain structures \cite{zhou2025unified}. The model achieves a Dice Similarity Coefficient (DSC) of 0.813 for whole brain structures and 0.859 for tumor and demonstrates robust performance for important brain regions. This study provides physicians maximum benefit from the two proposed innovations by combining the unified segmentation model with high-quality uncertainty learning. The result is the first model that not only outputs the cancer in its natural surroundings but also with an uncertainty map overlaid on top. Visual investigation of numerous output images strongly suggests the uncertainty information provides critical insights to assess cancer prediction results and compensate for potential errors, a feature that would aid physicians in making more informed surgical planning decisions from AI predictions.

\newpage
\setcounter{page}{1}

\section{Introduction}
Brain tumors are among the most deadly neurological diseases. More than a million Americans suffer from brain tumors, with 100,000 new cases diagnosed each year, according to the Central Brain Tumor Registry of the United States (CBTRUS) \cite{BraTS2023}. Gliomas, the most frequent form of brain tumors, carry a poor prognosis. The median survival in patients with glioblastoma is less than 15 months, even under the most favorable treatments. Additionally, treatment poses an enormous economic burden. The accurate segmentation of brain tumors is essential for treatment, surgical resection, and radiotherapy. However, it remains largely inaccessible in routine clinical practice.

Currently, magnetic resonance imaging (MRI) is the dominant modality for brain tumor diagnosis because of its ability to image soft tissue with high contrast. In practice, multiple MRI modalities are merged—e.g., T1-weighted, T1-contrast enhanced (T1ce), T2-weighted, and fluid-attenuated inversion recovery (FLAIR)—to contour tumor subregions \cite{desikan2006}. Despite the detailed information provided by these combined sequences, accurate tumor segmentation remains a difficult task due to the heterogeneous nature of brain tumors. Different subregions, such as edema, necrotic core, and enhancing tumor tissue, exhibit different appearances across MRI sequences. Manual annotation requires radiologists and neurosurgeons to delineate tumor borders across modalities. This is a time-consuming, resource-intensive, and inter-observer variable process. Delineation errors can result in the inability to fully excise tumors, injure adjacent tissue, and have direct effects on patient outcomes.

Given these challenges, the development of accurate and reliable automated brain tumor segmentation methods has become a central focus of current medical imaging research. In particular, deep learning-based approaches have led to extraordinary progress in this field and are moving closer to clinical application \cite{fischl2012,shattuck2002}. In addition, transformer-based architectures and diffusion models have recently demonstrated improvements in understanding global tumor context. Among such approaches, the U-Net has emerged as a standard baseline \cite{ronneberger2015}. The architecture of the model encompasses both local and global context, and its automatic preprocessing, augmentation, and hyperparameter optimization render it extremely efficient on various datasets \cite{nnUnet}. As a result, U-Net based architectures have achieved DSC scores of 0.92 on the BraTS dataset.

Despite great advances, the discrepancy between benchmark performance and clinical usability remains a big challenge for real-world adoption. Current methods are evaluated almost exclusively by accuracy metrics such as the Dice Sørensen coefficient (DSC) or Intersection-over-Union (IoU) \cite{zhou2024metastasis}. Although these measurements reflect overall overlap between prediction and ground truth, they provide no insight into how confident the model is at the patient or voxel level. A model may achieve a high DSC but misclassify small but clinically significant regions, e.g., tumor margins near critical brain structures. Furthermore, deep learning models are known to be overconfident and prone to hallucinations. To clinicians, this results in a black-box effect: they are presented with a predicted mask but given no indications of which areas are trustworthy and which require meticulous human examination. In neurosurgical settings, millimeter-scale misclassifications have the potential to impact resection planning and patient outcome, and this lack of transparency is a major obstacle. Uncertainty quantification is a potential solution. By providing voxel-wise or region-level confidences, models can distinguish between high-confidence predictions and those that remain ambiguous. This allows clinicians to rely on portions where there exists high certainty and utilize their expertise on regions labeled as uncertain.

Equally important is the need for anatomically aware segmentation. Current tumor segmentation models often operate in isolation from surrounding brain structures, treating the tumor as a localized object rather than part of an interconnected anatomy. In neurosurgical planning, understanding the spatial relationship between tumors and critical functional regions is essential for preserving patient function. Our previously published unified model addresses this by jointly segmenting tumor and whole brain structures \cite{zhou2025unified}. Furthermore, most deep learning models are trained and validated on benchmark datasets such as BraTS, which represent only a narrow subset of scanner types, institutions, and patient populations. In practice, brain MRIs vary widely in acquisition protocols, magnetic field strengths, voxel resolutions, and patient demographics. As a result, models trained on one dataset often fail when exposed to new domains. Achieving robustness across multiple datasets, scanners, and brain types is critical for real-world usability. Future models must therefore be designed to handle heterogeneous data distributions and maintain consistent performance across diverse clinical settings.

\section{Related Works}
Recent work has sought to quantify model uncertainty to improve interpretability and trust in automated brain tumor segmentation \cite{mehta2022qubrats,jungo2020analyzing}. Sampling-based Bayesian approaches, such as Monte Carlo dropout and deep ensembles, estimate epistemic uncertainty through multiple stochastic forward passes and achieve strong correlations with segmentation error, but are computationally prohibitive for clinical use \cite{wang2019aleatoric}. Aleatoric uncertainty methods, including test-time augmentation and variance-predicting networks, capture data-driven variability more efficiently yet provide heuristic confidence estimates without formal guarantees \cite{wang2019aleatoric}. Calibration techniques such as temperature scaling and isotonic regression help align predicted probabilities with empirical accuracy but do not ensure correctness for individual predictions or localized voxel regions.

More recently, conformal prediction frameworks and evidential deep learning approaches have been proposed to generate voxel or region-level prediction sets with guaranteed statistical coverage or reliability learning \cite{li2023region,huang2024deep}. These methods introduce the appealing notion of finite-sample reliability but naïve implementations often yield overly conservative confidence maps that span large, clinically irrelevant regions. Even the most efficient conformal variants struggle to integrate spatial dependencies, attention-aware fusion, or achieve the inference speeds required for intraoperative use \cite{zhou2023uncertainty,yang2024spu}. Collectively, these approaches have advanced the theoretical foundations of uncertainty quantification but still lack a single-pass, spatially informed, and computationally efficient framework that provides interpretable confidence scores at clinically useful resolution.

For a comprehensive review of brain tumor segmentation and whole brain segmentation literature, including U-Net variants, DeepMedic, FCN, FreeSurfer, BrainSuite, SLANT, and FastSurfer, see our previously published work \cite{zhou2025unified}.

\subsection{Research Gap and Proposed Solution}
Despite advances, several research gaps persist across current methodologies for uncertainty prediction. Current methods rely heavily on inherent data noise in the training model and capture aleatoric components of uncertainty. Epistemic components, including segmentation errors, are not included. There is no current work in brain tumor segmentation that integrates segmentation labels, model uncertainty, and supervised training to improve uncertainty prediction to be anatomy-context aware. In addition, these techniques often require multiple network layers or multiple inference passes (e.g., Monte Carlo dropout and Deep Ensemble methods), which often cause minor decreases in cancer segmentation accuracy and require significantly more computational resources for training. This limits their use in clinical adoption.

Although recent advances have been made to segment the whole brain into cortical and subcortical structures, they have been applied only to healthy brain images. Providing a unified model to segment cancer and surrounding brain structures has been made difficult by the absence of datasets that contain labels for both cancer and other regions in cancer patients. Existing datasets are either for cancer labels (e.g., BraTS datasets) or for healthy patients with cortical/subcortical labels (e.g., OASIS-1), but not combined. However, models trained with high accuracy on healthy brains may not work well for patients with cancerous brains, since the brain anatomy is distorted and sometimes even altered by invasive cancer tissues. In addition, segmenting cancer and many healthy brain regions requires sophisticated models to provide accuracy for numerous complex image patterns.

Addressing this gap is essential for advancing segmentation models from experimental research to practical use in neurosurgical and radiotherapy workflows. Tumor segmentation models, while highly accurate, focus narrowly on pathological tissue and fail to account for the surrounding healthy anatomy, depriving surgeons of crucial spatial context needed to plan safe resection paths and preserve function. Due to a lack of available datasets, previous studies on whole brain segmentation of patients with tumors are scarce. It is suspected that distortion caused by brain tumors would make models trained on healthy brains alone less effective. This could have undesirable downstream consequences if such models were applied in clinical preoperative surgical use.

To address these research gaps, this work develops a novel uncertainty-aware model for brain tumor segmentation and uncertainty estimation. The proposed method extends a 3D nnU-Net with an architecture that predicts confidence for the model's produced segmentation cancer masks. The prediction is performed with a single additional network channel, with a loss function added to account specifically for uncertainty-related metrics. Building upon our previously published unified segmentation framework \cite{zhou2025unified}, this work develops the first model to segment both brain lesions and critical anatomical structures while quantifying prediction reliability at each voxel. Finally, the unified model was trained with the additional uncertainty prediction channel similar to the cancer-only prediction case, allowing physicians not only to analyze brain tumors in their natural surroundings but also to grasp an understanding of where the predictions are made with high confidence or suspicion.

\section{Methodology}
\subsection{Model Development}
\begin{figure}[h]
    \centering
    \includegraphics[width=0.85\textwidth]{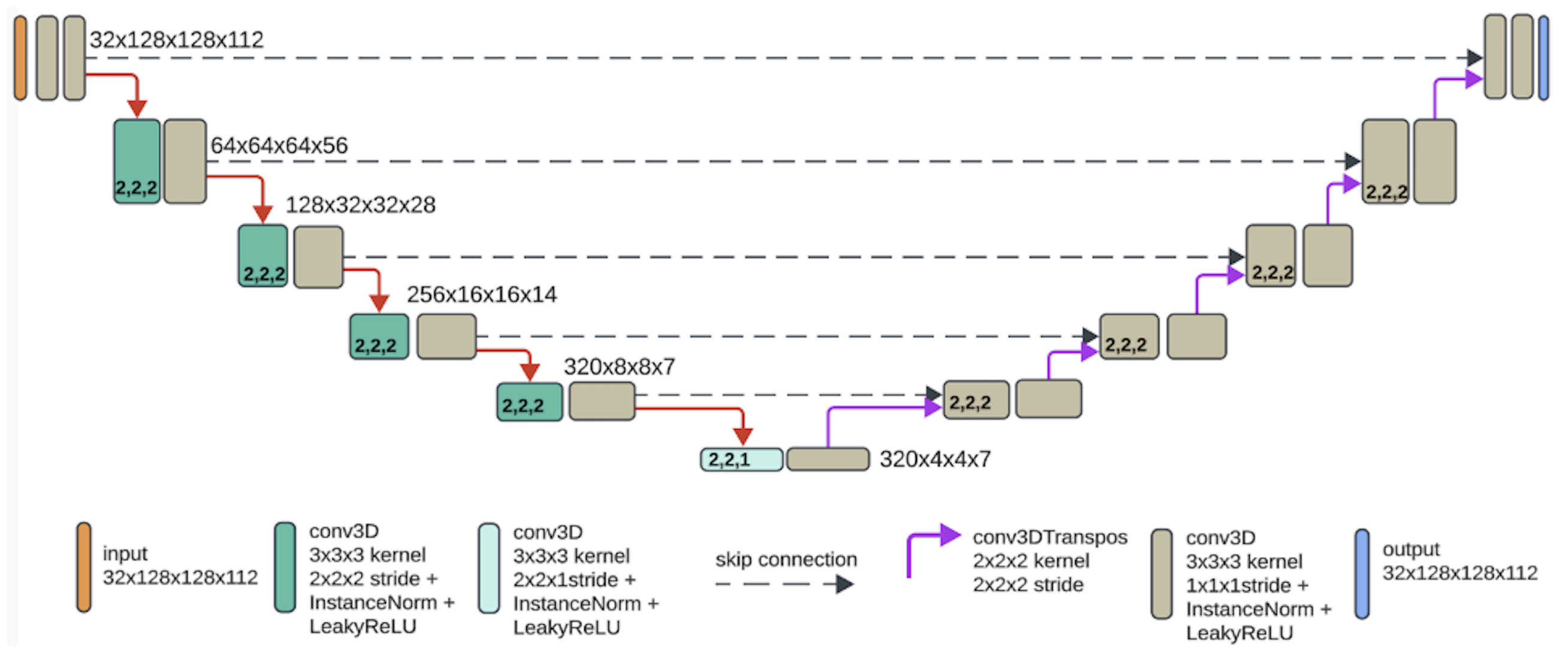}
    \caption{3D nnU-Net architecture.}
    \label{fig:nnunet_architecture}
\end{figure}

The present study utilizes the 3D nnU-Net model (Figure~\ref{fig:nnunet_architecture}) for all training procedures. Two types of prediction models were trained: cancer prediction only models (CMODEL) and unified prediction models (UMODEL) that predict both cancer and surrounding healthy regions. Additionally, modifications to the generic nnU-Net architecture were implemented to enable voxel-wise uncertainty prediction alongside segmentation results, as described in the sections below.

For each model type, two training runs were performed: one with an additional uncertainty channel and one without for comparison purposes. Therefore, four total training runs were conducted:
\begin{itemize}
    \item Cancer-only prediction using the BraTS2023 dataset, \textbf{CM1} with uncertainty prediction (for cancer only) and cancer segmentation using the combined loss function.
    \item Cancer-only prediction using the BraTS2023 dataset, \textbf{CM2} without uncertainty prediction using the default loss function (for cancer segmentation comparison).
    \item Unified model prediction using the BraTS2023 and preprocessed OASIS-1 datasets, \textbf{UM1} with uncertainty prediction (for cancer only) and cancer/whole brain segmentation using the combined loss function.
    \item Unified model prediction using the BraTS2023 and preprocessed OASIS-1 datasets, \textbf{UM2} without uncertainty prediction and cancer/whole brain segmentation using the default loss function.
\end{itemize}

The nnU-Net was trained following a standard configuration, using patch sizes of 128$\times$128$\times$128, batch sizes of 4, and an Adam optimizer. Learning rates decayed linearly from 0.001 to 0 over 500 epochs, and 125 training batches and 25 validation batches are done in each epoch. All training was conducted on NVIDIA A10 GPUs.

\subsection{Dataset}
This study employs the same datasets, preprocessing, and label reduction strategy as detailed in our previous work \cite{zhou2025unified}: BraTS2023 (1,248 patients) and OASIS-1 (425 subjects). BraTS2023 was preprocessed with FastSurfer to generate 52 whole brain labels; tumor masks were combined into a single label. For full details on data treatment, FastSurfer augmentation, and label mapping, see \cite{zhou2025unified}.

\subsection{Uncertainty Prediction Channel}
\begin{figure}[h]
    \centering
    \includegraphics[width=0.85\textwidth]{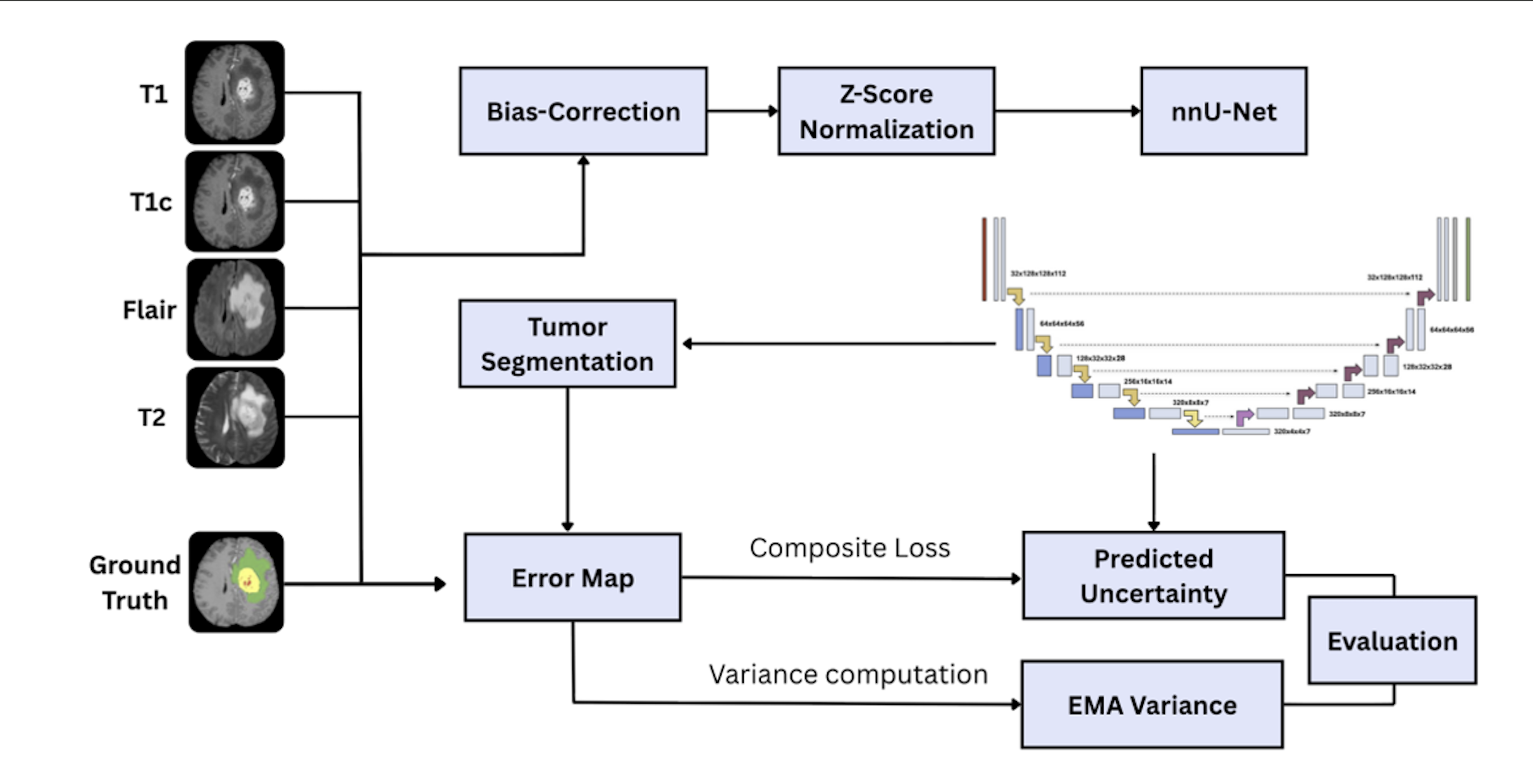}
    \caption{Proposed uncertainty prediction process.}
    \label{fig:unc_pred_process}
\end{figure}

In order to allow voxel-wise uncertainty prediction, the standard nnU-Net architecture is modified as shown in Figure~\ref{fig:unc_pred_process}. The modification begins with the computation of an error map for each training step, calculated as the difference between ground truth labels and predicted cancer labels. All subregions in cancer are combined before comparison so that the error only captures false positive or false negative cancer voxels with no distinction between different cancer regions. A dedicated uncertainty prediction channel is then added to the nnU-Net architecture alongside the standard segmentation output channels. This channel predicts uncertainty logits which are subsequently compared and optimized with the observed actual error map. Finally, an evaluation loss function is incorporated to compare the error map with uncertainty prediction, and this loss is added to the total loss for backpropagation.

It is noteworthy that the uncertainty channel operates independently, receiving gradients exclusively from uncertainty-specific loss components and not sharing information with the segmentation logits during training. This design offers several key advantages. First, it significantly reduces computational overhead. Second, it minimizes bias by preventing the uncertainty predictions from becoming dependent on segmentation results. Third, it prevents the uncertainty training from affecting cancer segmentation performance. The training objective for this channel is to maximize sensitivity to regions where the model's predictions are likely to be erroneous, thereby serving as a complementary quality assessment tool.

\subsection{Uncertainty Target Construction}
Training the CM1 and UM1 uncertainty channels uses the spatially averaged error map as the voxel-level uncertainty prediction target. At each epoch, a binary error map $X_t$ is generated by comparing the model's predicted combined cancer labels to the ground truth. Each voxel is assigned a value of 1 if the prediction differs from the ground truth and 0 otherwise. To reduce local noise, spatial smoothing is applied with a $3 \times 3 \times 3$ averaging filter:
\begin{equation}
\tilde{X}_{t} = \frac{1}{27}\sum_{i=-1}^{1}\sum_{j=-1}^{1}\sum_{k=-1}^{1} X_{i, j, k},
\end{equation}
where $X_t$ is the error map at epoch $t$. This is computed for the entire image and is not restricted to cancer voxels or brain regions.

\subsection{Loss Function}
The training loss function combines segmentation loss with RMSD loss and Pearson correlation between the predicted uncertainty and the uncertainty target. Let
\begin{itemize}
    \item $U$ be the uncertainty prediction output (as in the previous section),
    \item $E$ be the error map,
    \item $M$ be the mask of true cancer voxels.
\end{itemize}
Then,
\begin{equation}
\mathcal{L} = \mathcal{L}_{DCE} + \lambda_{RMSD} \mathcal{L}_{RMSD}+ \lambda_{corr} \mathcal{L}_{corr},
\end{equation}
where $\mathcal{L}_{DCE}$ is the hybrid Dice / Cross-Entropy loss used for the segmentation of the necrotic core, edema, and enhancing tumor regions;
\begin{equation}
\mathcal{L}_{RMSD} = \sqrt{\frac{\sum (U-E)^2M}{M + \epsilon}}
\end{equation}
is the RMSD between uncertainty target and true error; and
\begin{equation}
\mathcal{L}_{corr} = \frac{\sum M (U - \bar{U})(E-\bar{E})}{\sqrt{\sum M(U - \bar{U})^2 \sum (E - \bar{E})^2}}
\end{equation}
is the Pearson correlation between uncertainty target and true error. The weight hyperparameters were set to $\lambda_{RMSD} = 0.1$ and $\lambda_{corr} = 0.01$.

The combination of both RMSD and correlation losses provides complementary training signals for the uncertainty-aware model. The RMSD loss verifies that predictive uncertainty remains close to the observed error magnitudes. RMSD loss will penalize systematic underestimation or overestimation. In contrast, correlation loss ensures regions of anticipated high uncertainty coincide with regions where segmentation errors are most likely to occur. RMSD in isolation would develop accurate calibrations but may not effectively recover important spatial patterns; correlation loss alone can capture the shape of error but ignore absolute accuracy. The combination of these two factors enables the model to produce confidence maps that are accurate and reliable at voxel and case level scales.

\subsection{Evaluation Metrics}
To evaluate the proposed uncertainty-aware model, this work tested both segmentation accuracy and uncertainty estimation quality. Evaluation was conducted on the validation set (20\% of the dataset) for each fold, with metrics averaged over the final 20 epochs.

Dice coefficient (DSC) is a common measure to present the congruence of predicted tumor regions with actual ground truth labels. Dice score is computed separately for each subregion of the tumor according to the formula:
\begin{equation}
DSC = \frac{2|A \cap B|}{|A| + |B|}
\end{equation}
where $A$ refers to the set of predicted voxels and $B$ refers to the set of ground-truth voxels. A score closer to 1 indicates improved segmentation precision. This measure is generally applied in BraTS benchmarks and provides easy comparability with previous research.

In addition to segmentation accuracy, the alignment of predicted uncertainty maps with true segmentation errors was evaluated by using the RMSD and Pearson correlations as described in Equations~(3) and~(4). The RMSD measures voxel-level differences, while the Pearson correlation assesses global similarity between predicted and true uncertainty levels.

\section{Results and Discussion}
In this section, the study first evaluates the performance of the cancer-only segmentation models and whole-brain segmentation models. Then we evaluate DSC for finer brain regions to demonstrate that the whole-brain model provides accurate segmentation on important regions. Finally, we examine the uncertainty plots in wholly segmented brains to show how our model could provide valuable insights for clinical practice.

\subsection{Model Uncertainty Performance}
Table~\ref{tab:combined_results_reordered} reports the cancer segmentation DSC for the different brain tumor regions and the uncertainty performance metric for the CM and UM models. Table~\ref{tab:uq_corrected} compares the uncertainty metric against studies in published literature.

\begin{table}[htbp]
\centering
\caption{Performance comparison for segmentation and uncertainty for all models. Dice scores are reported as \textbf{DSC(region)}, Uncertainty metric are reported as: RMSD as \textbf{UNC(rmsd)} and CORR as \textbf{UNC(corr)}. N/A = not applicable. Table created by the student researcher, 2025.}
\label{tab:combined_results_reordered}
\begin{tabular}{lcc|cc}
\toprule
\textbf{Region / Metric} & \textbf{CM1} & \textbf{CM2} & \textbf{UM1} & \textbf{UM2} \\
\midrule
\multicolumn{5}{c}{\textit{Cancer Segment}} \\
\midrule
\textbf{DSC(Whole tumor)} & 0.909 & 0.910 & — & — \\
\textbf{DSC(Tumor core)} & 0.910 & 0.911 & — & — \\
\textbf{DSC(Enhancing tumor)} & 0.875 & 0.872 & — & — \\
\midrule
\textbf{DSC(Cortical)} & — & — & 0.790 & 0.793 \\
\textbf{DSC(Subcortical)} & — & — & 0.855 & 0.856 \\
\textbf{DSC(Whole brain)} & — & — & 0.813 & 0.815 \\
\textbf{DSC(Tumor)} & — & — & 0.859 & 0.861 \\
\midrule
\textbf{UNC(rmsd)} & 0.032 & N/A & 0.047 & N/A \\
\textbf{UNC(corr)} & 0.634 & N/A & 0.750 & N/A \\
\bottomrule
\end{tabular}
\end{table}

\begin{table}[htbp]
\centering
\caption{Uncertainty quantification in brain tumor segmentation using true per-voxel error (RMSD) and Pearson correlation coefficient (CORR). N/A = not reported. Table created by the student researcher, 2025.}
\label{tab:uq_corrected}
\begin{tabular}{l cc}
\toprule
\textbf{Method} & \textbf{RMSD} & \textbf{CORR} \\
\midrule
\multicolumn{3}{c}{\textit{Our Models}} \\
\midrule
\textbf{CM1 (Ours)} & \textbf{0.032} & 0.634 \\
UM1 (Ours) & 0.047 & \textbf{0.750} \\
\midrule
\multicolumn{3}{c}{\textit{Literature}} \\
\midrule
Wang (2019)~\cite{wang2019aleatoric} & N/A & 0.71 \\
Jungo et al. (2020)~\cite{jungo2020analyzing} & 0.065 & 0.68 \\
Kohl et al. (2021)~\cite{savadikar2021} & 0.053 & 0.74 \\
Zhang et al. (2023)~\cite{zhang2023} & 0.046 & 0.70 \\
Nair et al. (2024)~\cite{mehta2024} & 0.059 & 0.64 \\
\bottomrule
\end{tabular}
\end{table}

Metrics demonstrate that the UM1 model excels at uncertainty prediction, as its Pearson correlation coefficient of 0.75 represents state-of-the-art performance compared to other studies (below). The CM1 model achieved a correlation coefficient of 0.634 for uncertainty prediction, which is lower than UM1 but still remarkable considering the model had to segment 54 different region labels. Interestingly, the voxel-wise uncertainty RMSD metrics reveal a different trend: UM1 achieved 0.047 while CM1 achieved 0.032, with CM1 actually performing better. In fact, the RMSD metric reported for CM1 is the lowest among the studies compared in Table~\ref{tab:uq_corrected}. The RMSD value provides a direct voxel-wise evaluation between predicted uncertainty and real error and indicates that our models, particularly CM1, performed better than existing statistics-based methods in localized uncertainty prediction.

\subsection{CM Models Uncertainty Prediction Visualization}
\begin{figure}[H]
    \centering
    \includegraphics[width=0.75\textwidth]{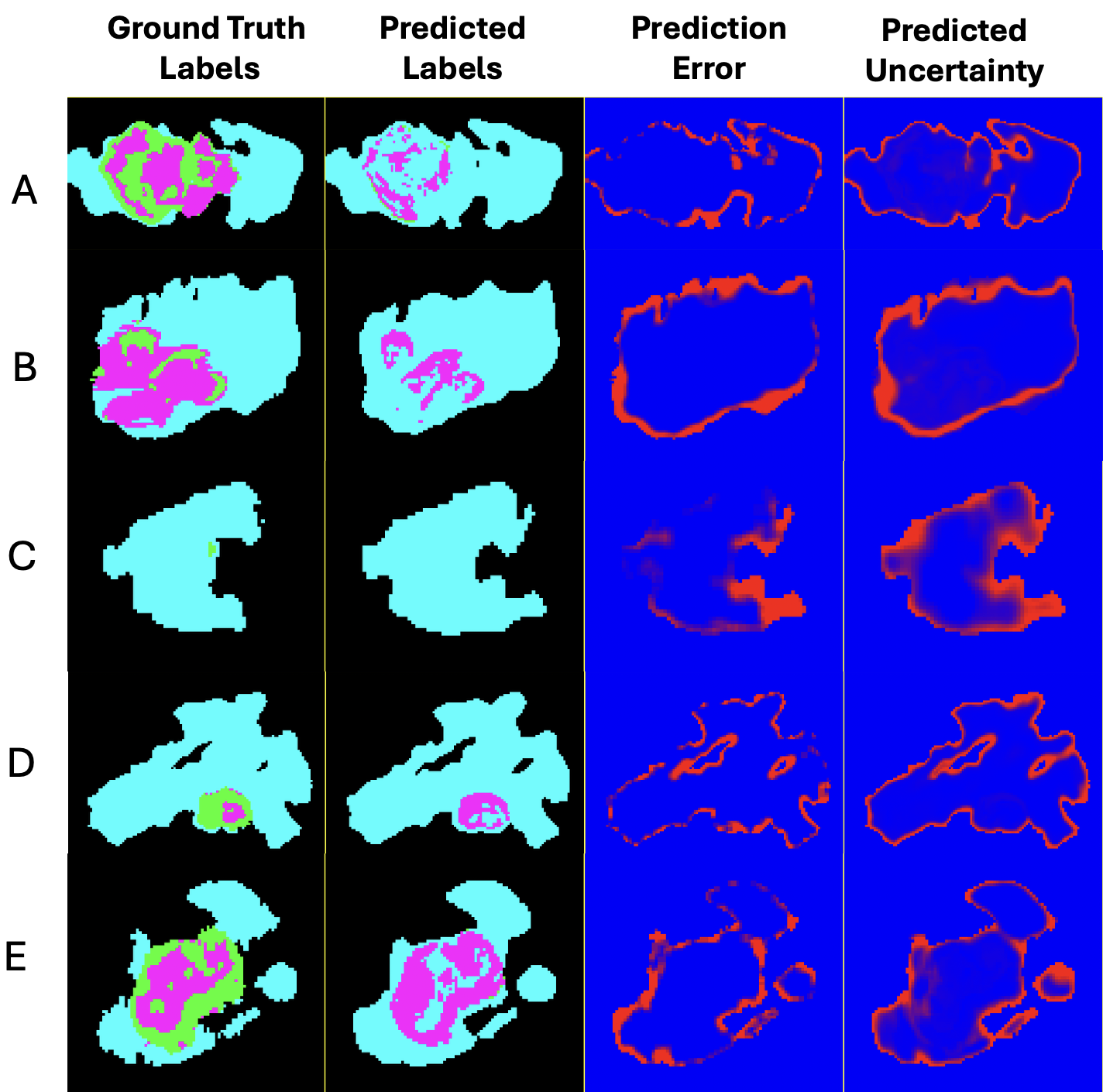}
    \caption{Visual comparison of CM1 segmentation and uncertainty prediction results. }
    \label{fig:cm1_unc_view}
\end{figure}

Figure~\ref{fig:cm1_unc_view} displays several samples with the ground truth labels, predicted cancer labels, prediction error, and predicted uncertainty maps. For ground truth and predicted cancer labels, the whole tumor is in cyan, tumor core in purple, and ET in green. For error and uncertainty maps, uncertain regions are in red and certain regions are in blue.

We may use plots such as Figure~\ref{fig:cm1_unc_view} to qualitatively assess segmentation and uncertainty predictions. For segmentations, the whole tumor prediction closely matches the ground truth, and the general shape of the tumor core is often detected, but there are some examples of ET regions missing from the prediction or even being mistaken for the core. However, the overall tumor contour is often accurately determined, as evidenced by the higher DSC of the whole tumor/tumor core.

Unsurprisingly, much of the error is concentrated around the whole tumor boundary. The uncertainty map captures the high ambiguity at the boundary but will also sometimes express slight uncertainty for a wider region around it. This may possibly be attributed to the smoothing of the error map fed into the uncertainty head, but on closer inspection, these regions sometimes also trace out boundaries between ET and tumor core.

\subsection{Model Segmentation Performance}
While the addition of a new channel aids uncertainty prediction, it does not appear to adversely affect cancer prediction accuracy. Not only is the segmentation DSC of our CM1 model close to that of CM2, but it is also typical of other 3D nnU-Net models trained on BraTS using all four modalities of input MRI images. In recent studies, DSC for whole tumor, tumor core, and enhancing tumors have been reported in the ranges of 0.90--0.92, 0.89--0.91, and 0.88--0.91. The scores of CM1 represent above-average performance for tumor core and only slightly worse performance for ET regions, demonstrating that it is comparable with state-of-the-art work.

Augmenting the model to segment normal brain structures alongside cancer has only a slight detrimental effect on cancer segmentation quality. While the UM2 model's tumor segmentation DSC of 0.86 is lower than that of CM2 (at 0.91), it still performs respectably well, especially considering that the model must segment 54 labels instead of only three cancer labels. Adding an uncertainty head on top of a unified model (UM1) also does not affect cancer segmentation accuracy, with its tumor segmentation DSC of 0.859 nearly identical to that of UM2 (at 0.861). As for healthy regions, it achieves a DSC of 0.813 for the whole brain, 0.855 for subcortical regions, and a slightly lower value of 0.79 for cortical regions, showing that it can handle segmentation of other brain regions reasonably well. These results are consistent with our previously reported unified model performance \cite{zhou2025unified}.

\subsection{DSC of Individual Brain Regions}
\begin{figure}[H]
    \centering
    \includegraphics[width=0.85\textwidth]{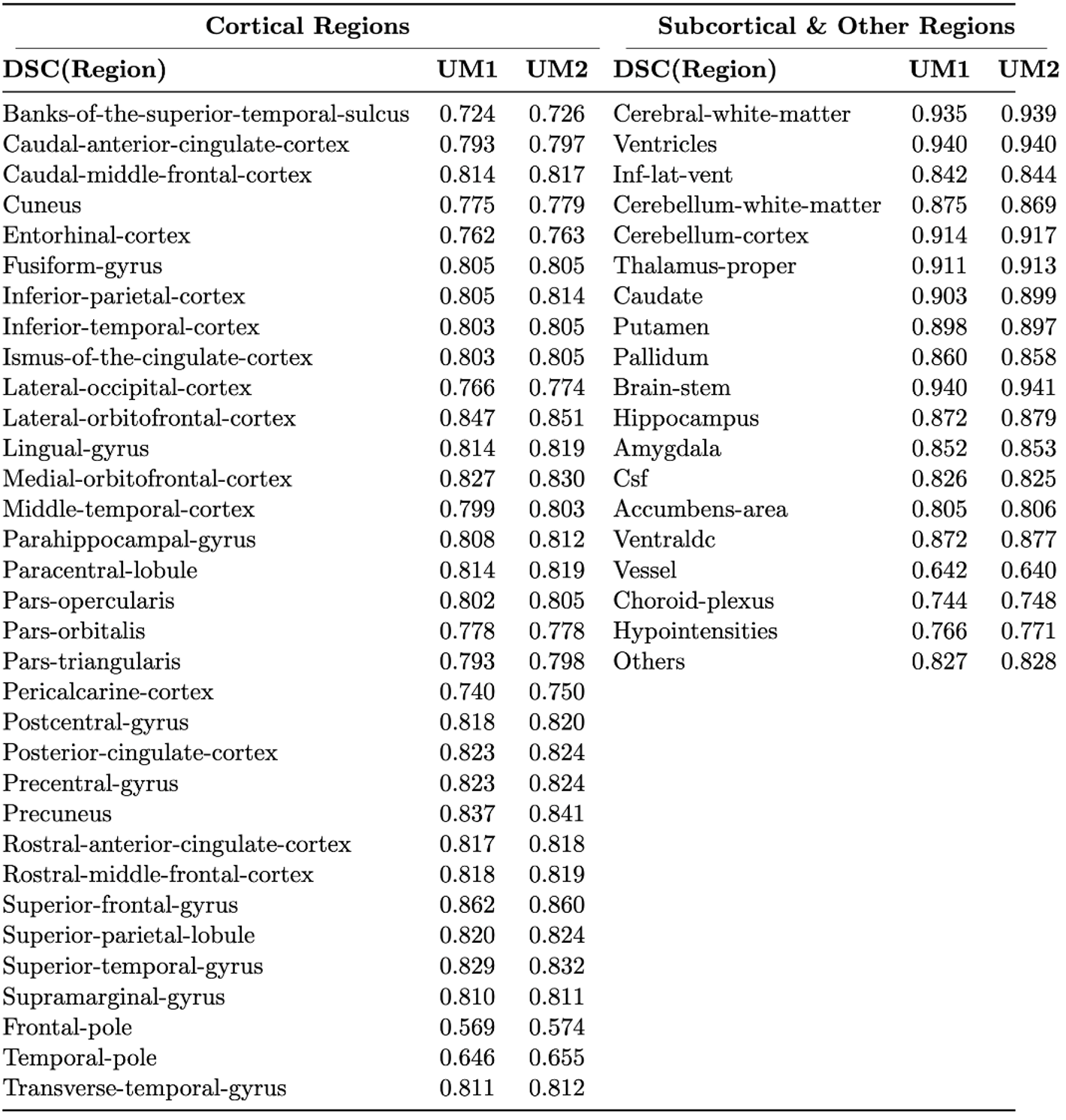}
    \caption{Comparison of DSC of detailed brain regions in unified models. }
    \label{fig:dsc_53labels}
\end{figure}

While achieving all-around segmentation performance is desirable from an engineering perspective, certain brain regions are more critical for function than others and may require greater segmentation accuracy for surgical purposes. For this reason, a detailed examination of particular regions may prove valuable. Figure~\ref{fig:dsc_53labels} lists the DSC of all whole brain segmentation labels, including both cortical and subcortical regions. Overall, UM1 and UM2 gave similar segmentation results.

A few regions are of particular interest to physicians. Brain stem segmentations are valued because they are useful for detecting brainstem compression or infarcts, so that even small errors could misguide interventions. Here the unified models achieved a high DSC of 0.94. Ventricle segmentations are crucial for diagnosing ventricular enlargement in hydrocephalus or atrophy-related conditions. Again, our models attained a high DSC of 0.94. Hippocampus segmentation must be unusually accurate due to the small, convoluted shape of this key structure. Our models attained a DSC of near 0.87.

Although segmentation of either cancer or whole brain structures has been done in many previous studies, segmentation for cancer together with healthy regions in the same AI model has not been done before. Our models suggest this is not only possible but probably the desirable approach for clinically relevant output.

\subsection{Unified Segmentation with Uncertainty}
In order to use the information obtained from the unified segmentation models (both cancer and surrounding regions) and the information obtained from the uncertainty prediction channel to aid physicians, a program was written to plot the segmentation data into a graph with uncertainty data overlaid. In this graph, all 53 normal regions are displayed with different colors, while the tumor is displayed in yellow with a red hue overlaid using the uncertainty output for each voxel. The intensity of the red hue is proportional to the uncertainty levels: no overlay if uncertainty is zero, maximum intensity when uncertainty is 1. This provides physicians the unique experience to view tumors surrounded by important brain regions and know where to check for problems (as indicated by high uncertainty levels).

The main benefits of this approach are threefold. First, combining tumor and whole brain information into one graph enables visualization of interaction between tumor and its surroundings. Second, combining segmentation data and uncertainty enables viewing not just prediction but also prediction confidence. Third, to the best of our knowledge, this is the first time this kind of map has been created to provide physicians a holistic view of AI-predicted results.

Using this approach, we analyzed approximately 50 patient images and summarize the findings by grouping them into three main cases below:

\subsubsection{Case 1: Prediction Matches Reality (Uncertainty Highlights Boundary)}
\begin{figure}[h]
    \centering
    \includegraphics[width=0.65\textwidth]{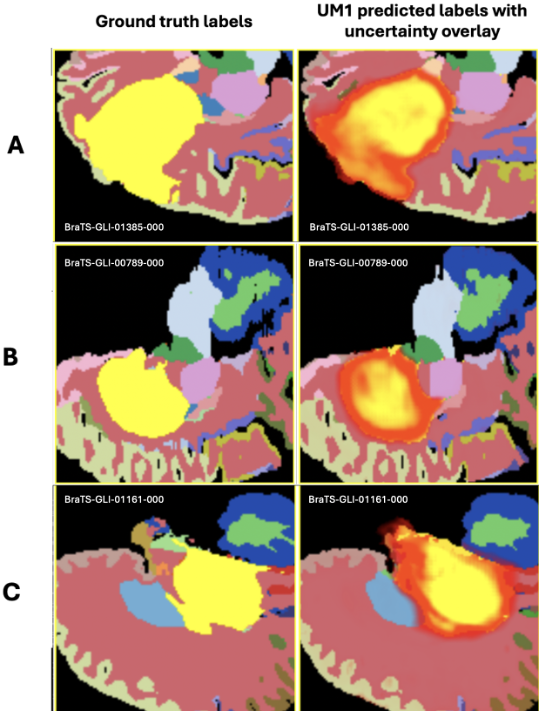}
    \caption{Case 1: Prediction matches reality with uncertainty highlighting boundary. }
    \label{fig:case_1}
\end{figure}

Figure~\ref{fig:case_1} shows a group of images where the predicted cancer labels mostly align with ground truth labels, while the uncertainty prediction naturally lies at the tumor boundary and highlights the less confident areas surrounding the boundary. The images are shown in the area adjacent to the cancer.

\subsubsection{Case 2: Prediction Over-predicts (Uncertainty Catches Suspicious Regions)}
\begin{figure}[h]
    \centering
    \includegraphics[width=0.65\textwidth]{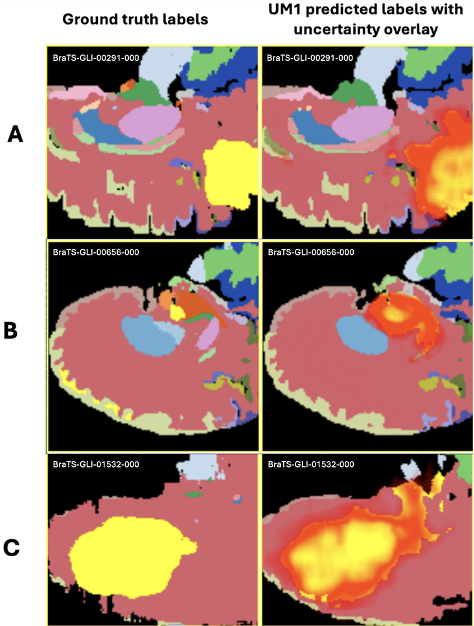}
    \caption{Case 2: Prediction over-predicts with uncertainty catching suspicious regions. }
    \label{fig:case_2}
\end{figure}

Figure~\ref{fig:case_2} shows a group of images where the predicted cancer labels cover more area than the ground truth labels, signifying overprediction by the UM1 model. The uncertainty predictions are very likely to coincide with the over predicted region, which provides helpful hints for physicians to manually investigate these regions. Patient A: cancer is on the right side of the brain, and the model over predicts cancer and marks a large area in the middle of the brain as cancerous. The uncertainty overlay captures most of the overpredicted area. Patient B: a small cancer on the top part of the brain where the model over predicts the size of the cancer. The uncertainty overlay captures the over predicted outer area of the tumor, while the original cancer in the middle shows less uncertainty. Patient C: tumor exists on the left side of the brain, and the model over predicts on the right side. The uncertainty overlay highlights the over predicted area on the right.

\subsubsection{Case 3: Prediction Under-predicts (Uncertainty Catches Potential Misses)}
\begin{figure}[h]
    \centering
    \includegraphics[width=0.65\textwidth]{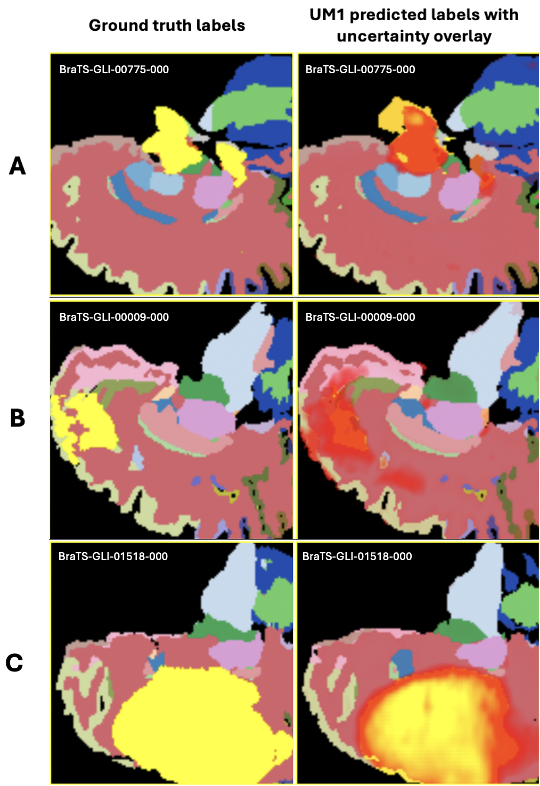}
    \caption{Case 3: Prediction under-predicts with uncertainty catching potential misses. }
    \label{fig:case_3}
\end{figure}

Figure~\ref{fig:case_3} shows a group of images where the predicted cancer labels cover less area than the ground truth labels, signifying under prediction by the UM1 model. The uncertainty predictions are also likely to coincide with the missed cancer region, which provides helpful hints for physicians to manually investigate these regions. Patient A: cancer is in the top part of the brain, and the model misses one of the cancer regions on the right and under predicts the region on the left. The uncertainty overlay captures both misses for further investigation. Patient B: cancer is on the left side of the brain, and the model significantly under predicts. However, the uncertainty overlay highlights the wedge-shaped area that closely resembles the real cancer that was missed. It also highlights another suspicious area to the right of it. Patient C: large tumor mass on the bottom side, but the prediction misses the tip on the right side, and the uncertainty overlay captures that part again.

\subsection{Limitations of Current Research}
There are several main limitations that restricted further improvement of this study. The unified model used only one modality instead of all four MRI modalities available in BraTS (to be compatible with OASIS-1) due to limitations in computing resources available at home and time constraints. This can lead to reduced accuracy in the UM1 and UM2 models. We expect our model performance would further improve if it were able to incorporate more information with additional modalities. Also, the sample size of our healthy brain dataset (OASIS-1) is limited, with the possible effect that our segmentation of healthy regions is slightly less accurate than that for tumors. This issue could be remediated with access to larger private datasets, such as ADNI or OASIS-3. In addition, we used only a single uncertainty channel for uncertainty prediction. While our models performed well and this might be sufficient, it would be of interest to explore using more channels for the unified model uncertainty training, as there are 54 labels in the segmentation that might affect the uncertainty output. Most of these limitations can be addressed with access to additional databases, computing resources, time, and effort, and the scope of this research could certainly be expanded in future studies.

\section{Conclusion}
This study presents a novel approach to brain tumor segmentation that integrates uncertainty prediction with comprehensive anatomical segmentation. The proposed uncertainty prediction using additional channels works well in both CM1 and UM1 training, achieving state-of-the-art performance without affecting cancer segmentation or requiring excessive training resources. This allows the model to predict uncertainty in one pass instead of multiple inferences, which reduces training time and inference time. It also allows direct learning of uncertainty from the neural network using raw image data. This is an improvement over existing methods and provides a direction for future studies which might further enhance its accuracy.

Furthermore, building upon our previously published unified model \cite{zhou2025unified}, the unified segmentation for both tumor and healthy brain regions using combined real and synthetic data achieves robust results, especially for important subcortical regions. Visual inspection shows good correspondence between segmented labels and ground truth. This allows physicians to view the predicted cancer with context to its anatomical environment.

Combining these two techniques allowed the construction of a visual diagram of the cancer, its surrounding environment, overlaid with the uncertainty information. Examination of many patient cases reveals that such information is very helpful to doctors in making decisions for surgical planning, with the uncertainty information often highlighting areas of ambiguity that the doctor should pay attention to.

\bibliographystyle{ieeetr}
\bibliography{references}
\newpage
\appendix
\end{document}